# Rethinking 3D Segmentation from Individual LiDAR Scans: Incidence-Aware Sampling on the SIP Benchmark

Seongyong Kim[1], Jingdao Chen[2], and Yong Kwon Cho[3]

[1]Graduate Student, School of Civil and Environmental Engineering, Georgia Institute of Technology, 790 Atlantic Dr., Atlanta GA, E-mail: skim3310@gatech.edu

[2]Assistant Professor, Department of Computer Science and Engineering, Mississippi State University, 665 George Perry St, Mississippi State, MS 39762, E-mail: chenjingdao@cse.msstate.edu

[3]Professor, School of Civil and Environmental Engineering, Georgia Institute of Technology, 790 Atlantic Dr., Atlanta, GA, E-mail: yong.cho@ce.gatech.edu (Corresponding Author)

## ABSTRACT

3D scene understanding is increasingly important for construction applications, yet most existing methods are developed on curated datasets that do not fully reflect site sensing conditions. In many construction workflows, individual LiDAR scans serve as local site evidence for rapid updates rather than complete representations of the scene. The resulting point clouds contain limited surface coverage, acquisition-driven density variation, and strong imbalance between dominant planar surfaces and sparse construction elements. Since large point clouds must be downsampled, sampling resolution and point allocation determine how limited input capacity is distributed between geometric detail and spatial context. This study evaluates these effects under a fixed per-fragment point budget and introduces an incidence-aware sampling strategy for individual LiDAR scans. The proposed method transforms points into a geometry-normalized manifold space for voxel-based selection, while retaining their original Euclidean coordinates for downstream learning. It requires only point coordinates and normals and does not require modification of the downstream backbone, although its effectiveness depends on the backbone representation. The Site in Pieces (SIP) construction benchmark provides realistic individual-scan sensing conditions for controlled evaluation across sampling resolutions. Experiments with Point Transformer and PointNeXt show that the proposed sampling improves resolution-averaged segmentation performance, with larger gains for non-planar elements and particularly strong improvements for ladders. The method also reduces sensitivity to sampling resolution, indicating that acquisition-aware point allocation provides a more stable geometric representation to the models. These findings position sampling as an active component of individual-scan 3D segmentation rather than as a generic preprocessing step.

## Introduction

Construction is a long-established industry that continuously generates large volumes of data across the project lifecycle. These data span diverse sources, including design documents, schedules, cost records, inspection logs, and increasingly, sensor-based measurements [1], [2]. At the same time, the industry is undergoing a broader transition through the adoption of automation, robotics, and digital twin technologies, creating strong demand for a reliable data-driven understanding of the physical site [3], [4].

Among these data sources, 3D sensing provides the most direct representation of as-is site conditions. Robust 3D scene understanding is therefore a key capability for practical applications such as progress monitoring [5], [6], deviation detection [7], digital twinning [8], and risk-aware site assessment [9]. Recent advances in 3D perception have been driven by deep learning models trained on increasingly large and curated datasets, with growing emphasis on scalable backbones and general-purpose representations [10]. However, many of these advances are developed and validated under settings that differ substantially from real construction data acquisition.

In practice, construction sensing does not always aim for complete multi-view reconstruction at every update cycle. Instead, many workflows require rapid interpretation of local site conditions from a fixed scanner position or a small number of local viewpoints. This is a common practical condition on active sites, where occlusion, limited access, and changing layouts make full coverage difficult. As a result, the captured point clouds often contain incomplete surface observations and non-uniform sampling patterns [11].

Long-tailed category distributions are a recurring challenge in 3D scene understanding [12]. In construction scenes, this issue is further amplified because less frequent but operationally important elements often appear as thin, fragmented, or weakly observed structures. Under these conditions, the challenge is not only class imbalance in the label distribution, but also the fragility of the geometric evidence itself.

This observation makes sampling a central issue rather than a neutral preprocessing step. Conventional downsampling reduces point count in Euclidean space, but in single-scan LiDAR it can disproportionately suppress the limited evidence associated with thin or weakly observed structures while preserving redundant samples on dominant planar regions. The resulting representation bias is shaped jointly by acquisition geometry and sampling design. Because large point clouds must still be downsampled to fit practical GPU memory limits, this issue becomes further entangled with scale effects, as sampling resolution directly controls the trade-off between geometric detail and spatial coverage. Unlike 2D models that operate on fixed rasterized inputs, point-based 3D models allow the input density itself to be redistributed. This makes sampling not only a compression step, but also a mechanism for allocating limited input capacity toward geometrically informative regions.

This work addresses the problem by reframing sampling as an acquisition-aware front end for individual LiDAR scans. Rather than replacing modern 3D backbones, the proposed method performs lightweight geometric normalization before feature learning. It requires only point coordinates and normals, does not depend on labels, and preserves the original Euclidean coordinates of the selected points. The method does not require modification of the downstream backbone, although its effectiveness depends on how the backbone represents spatial neighborhoods.

To this end, an incidence-aware manifold sampling strategy is introduced to redistribute point density according to surface geometry while preserving the original spatial structure (Figure 1). The method is evaluated across sampling resolutions and hierarchical point-based backbones under a fixed per-fragment point budget. To support this analysis, the Site in Pieces (SIP) benchmark [13] provides individual terrestrial LiDAR scans that preserve limited surface coverage, acquisition-driven density variation, and severe class imbalance in active construction environments. The main contributions of this work are summarized as follows:

- An incidence-aware manifold sampling strategy that redistributes points according to acquisition geometry while preserving their original Euclidean coordinates.
- A fixed-point-budget evaluation protocol for examining sampling behavior across resolutions and hierarchical point-based backbones.
- Consistent resolution-averaged performance gains across Point Transformer and PointNeXt, with the largest improvements observed for thin and weakly represented elements.
- The SIP benchmark, a construction-focused dataset that enables controlled evaluation under realistic individual-scan LiDAR conditions.

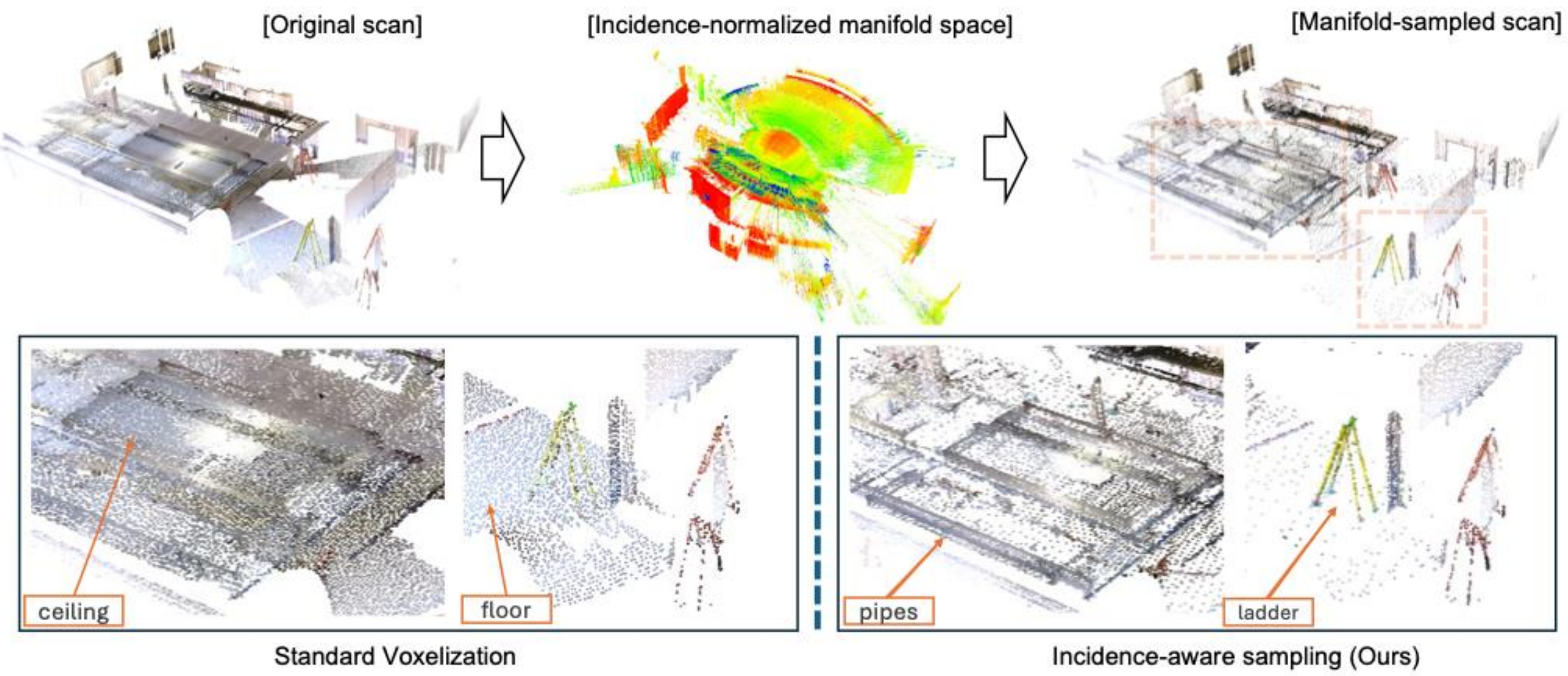

**Fig. 1.** Incidence-aware sampling for 3D training from single-scan LiDAR. Points are sampled to promote consistent coverage over the surface manifold, improving the balance of geometric cues. Compared to standard voxelization, planar regions are reduced while structurally thin regions are preserved. Color encodes normalized incidence, with red indicating grazing angles and blue indicating near-orthogonal surfaces.

## 2. Related Work

### *2.1 Point Cloud Sampling Methods*

Point cloud sampling is a fundamental component of 3D processing pipelines, serving to reduce computational cost while retaining useful geometric information. Classical strategies include random sampling, farthest-point sampling, and voxel- or grid-based sampling, which remain widely used for their simplicity and efficiency. Geometry-aware variants further attempt to preserve local structure by prioritizing curvature, edges, or other salient regions [14], [15], [16]. Taken together, these approaches are generally designed to compress point sets while maintaining geometric fidelity in Euclidean space.

More recently, sampling and point selection have increasingly been coupled with learned representations [17]. Instead of treating sampling as a fixed preprocessing step, recent methods often integrate point reduction, token selection, or adaptive pooling into the network itself, allowing the model to learn which subsets of the input are most useful for the target task [18], [19], [20]. While effective, these approaches are often task-specific and are not always directly aligned with point-wise semantic segmentation settings. This trend reflects a broader shift in 3D vision from hand-crafted sampling heuristics toward task-adaptive and representation-aware processing.

Despite these advances, most sampling strategies are not explicitly designed for the acquisition characteristics of single-scan LiDAR. In construction scenes, sampling irregularity is not only a matter of spatial density but also a consequence of partial visibility and strong structural redundancy on dominant surfaces. This motivates a front-end normalization perspective in which sampling is used to regularize acquisition-driven bias before feature learning, while remaining separable from downstream feature learning.

### *2.2 3D Datasets for Semantic Understanding*

The development of 3D scene understanding has been supported by a wide range of datasets across indoor, outdoor, and object-level settings. Indoor scene benchmarks such as S3DIS [21], ScanNet [22], ScanNet++ [23], Matterport3D [24], Structured3D [25], and ARKitScenes [26] have played a central role in semantic segmentation and scene parsing. These datasets provide rich supervision and large-scale benchmarks, but they often emphasize indoor environments captured through dense reconstruction or multi-view aggregation.

In parallel, large-scale LiDAR datasets from autonomous driving, including SemanticKITTI [27], nuScenes [28], and Waymo [29], have advanced outdoor 3D perception under real sensing conditions. These datasets are especially important in showing how scan geometry, sparsity, and viewpoint affect semantic understanding. However, their sensing setups, scene layouts, object distributions, and operational scales differ substantially from those of terrestrial scanning in construction environments.

Construction-domain 3D datasets remain comparatively limited. Existing efforts demonstrate the growing interest in site-specific perception, but they often differ in sensing setup, task definition, and reconstruction assumptions. Some focus on temporal registration or reconstruction-oriented problems [30], while others do not directly reflect the partial, single-scan conditions encountered in semantic segmentation workflows on active sites. This leaves a gap for benchmarks that explicitly preserve the sensing constraints of single-scan terrestrial LiDAR in cluttered, in-progress construction environments.

Within this context, SIP [13] is positioned not simply as another domain dataset, but as a benchmark designed to expose the interaction between acquisition-driven density variation, partial geometry, class imbalance, and scale-dependent sampling behavior.

### *2.3 Trends in 3D Deep Learning*

3D deep learning has evolved rapidly from early point-based networks toward a broader ecosystem of sparse convolutional models [31], [32], transformer-based architectures [33], [34], and hybrid representations [35], [36]. Point-based networks established the feasibility of direct point cloud learning, while later sparse convolution backbones improved scalability for large scenes. More recent transformer-based models further expanded the ability to capture long-range geometric relationships and hierarchical context, contributing to strong progress in 3D semantic understanding.

At the same time, the field has been moving beyond task-specific supervised models toward large-scale pretraining and more general 3D representations. Self-supervised learning [37], [38], masked modeling [39], cross-modal learning [40], and multi-dataset training [41] have all become increasingly prominent. This shift suggests that future 3D systems will rely not only on stronger backbones but also on more transferable representations learned across diverse sensing conditions and scene types.

In this landscape, the proposed method is intended as a complementary sampling front end rather than a replacement for modern 3D models. Because it requires only coordinates and normals, does not depend on labels, and retains the selected points in their original Euclidean coordinates, it can be incorporated before downstream feature learning. Its role is to regularize acquisition-driven density bias at the sampling stage so that the model receives a more balanced geometric representation.

## 3. Incidence-Aware Sampling via Surface Manifolds

### *3.1. Geometry Intuition: Sensor-Centered Sampling Bias*

Single-scan LiDAR point clouds do not represent volumetric occupancy. Instead, they consist only of points sampled from surfaces visible to the sensor, making the resulting distribution inherently sensor-dependent and non-uniform. This non-uniformity is not random, but a structured consequence of the acquisition geometry.

In particular, the density of observed points is shaped jointly by surface orientation and distance from the sensor. To describe the orientation effect, let $x \in \mathbb{R}^3$ denote a point, $n(x)$ the unit surface normal, and $v(x) = x/\| x \|$ the viewing direction from the sensor center. We define the incidence term as

$$\gamma(x) = | n(x)^\top v(x) |.$$

When $\gamma(x)$ is close to 1, the surface is viewed near orthogonally, whereas small $\gamma(x)$ corresponds to grazing incidence. Thus, even similar point spacing may correspond to different amounts of observed surface support. In addition, because LiDAR measurements are emitted in angular directions, the surface area covered by a fixed angular aperture increases with distance, so point density tends to decay approximately with range.

Together, these effects show that point density in single-scan LiDAR is governed by sensor-centered geometry. Accordingly, the goal of resampling is to promote more consistent support over the observed surface by accounting for local incidence and radial spread. This motivates the sensor-normalized manifold mapping introduced next, which adjusts sampling coordinates while preserving the original Euclidean coordinates for downstream learning.

*3.2. Manifold Mapping under Incidence-Induced Density Scaling*

Let $M \subset \mathbb{R}^3$ denote the observed surface manifold underlying a single-scan LiDAR point cloud. We define a geometry-normalized space $M'$ through the mapping

$$T: M \to M', x' = T(x) = x\sqrt{\gamma(x)}\frac{r_0}{r(x)},$$

where $r_0$ is a reference range and $\gamma(x)$ is the incidence term from Section 3.1. This mapping does not reconstruct the surface explicitly. Instead, it reparameterizes the observed points so that acquisition-driven density variation is reduced before sampling.

Because learning requires a finite set of representative points at each step, this continuous space must ultimately be discretized. The role of $M'$ is therefore not geometric reconstruction, but to provide a coordinate space in which discrete partitioning better reflects surface support under the sensing geometry.

Define

$$w(x) = \gamma(x)\left(\frac{r_0}{r(x)}\right)^2.$$

**Proposition.** Let $x \in M$, and assume that $M$ is locally smooth near $x$. If $\gamma(x)$ and $r(x)$ vary smoothly over the local tangent plane $T_xM$, then a local partition of $M'$ promotes a more consistent correspondence between occupied cells and represented surface support, up to first-order variation of $\gamma(x)$ and $r(x)$ within the neighborhood.

**Proof.** In a sufficiently small neighborhood of $x$, local smoothness allows $\gamma(x)$ and $r(x)$ to be treated as approximately constant on $T_xM$. The mapping is then locally equivalent to a scaling on the tangent plane with factor

$$c(x) = \sqrt{\gamma(x)}\frac{r_0}{r(x)}.$$

Thus,

$$J_T \mid_{T_xM} \approx c(x)\, I_2,$$

and the induced area element satisfies

$$dA' \approx \mid \det\left(J_T \mid_{T_xM}\right) \mid dA \approx c(x)^2 dA = w(x)\, dA.$$

Equivalently,

$$dA \approx \frac{dA'}{w(x)}.$$

Thus, the transformed space preserves local surface neighborhoods while reweighting the area measure by sensing geometry. Partitioning in $M'$ therefore reduces the tendency to over-represent grazing or near-range regions and yields a more consistent relation between occupied cells and surface support than direct partitioning in the original scan space.

### *3.3 Practical Sampling via Manifold Voxelization*

Rather than reconstructing the surface explicitly, the proposed method samples points by voxelizing the transformed manifold coordinates. This converts the continuous manifold coordinates into a finite set of discrete bins, providing an efficient front-end sampling scheme for 3D perception pipelines.

The practical transformation is defined as

$$x' = x\sqrt{\gamma(x)}\left(\frac{r_0}{r(x)}\right)^{\beta}$$

where $\beta$ controls the strength of radial normalization. When $\beta = 1$, the transformation follows the ideal sensor-normalized mapping described in Section 3.2. Smaller values of $\beta$ reduce the influence of range normalization, allowing the method to avoid excessive radial compression in sparse or distant regions. The target resolution h is interpreted as the nominal sampling resolution at the reference range $r_0$.

After this coordinate adjustment, standard voxel-based sampling is applied in the transformed space, and one representative point is retained from each occupied cell, as summarized in Algorithm 1.

**Algorithm 1.** Incidence-aware manifold sampling.

---

**Input:** point coordinates $X \in \mathbb{R}^{N\times 3}$, unit normals $N \in \mathbb{R}^{N\times 3}$, target surface resolution $h$, reference range $r_0$, radial exponent $\beta$
**Output:** selected point indices $\mathcal{I}$

1: $\mathrm{R} \leftarrow \| X \|_2$ , $V \leftarrow X/\mathrm{R}$
2: $\gamma \leftarrow \max(|\langle N, V\rangle|, \epsilon)$
3: // Manifold transformation
4: $X' \leftarrow X \odot \sqrt{\gamma} \odot (r_0/\mathrm{R})^{\beta}$
5: // Voxel indexing in transformed coordinates
6: $G \leftarrow \lfloor X'/\mathrm{h} \rfloor$
7: $\mathrm{K} \leftarrow \mathrm{hash}(B_r, B_\theta, B_\phi)$, $\mathrm{J} \leftarrow \emptyset$
8: **for each** unique $k \in K$ **do**
9: $i \leftarrow \mathrm{select}(\{j \mid K_j = k\})$
10: $\mathcal{I} \leftarrow \mathcal{I} \cup \{i\}$
11: **return** $\mathcal{I}$

---

The transformed coordinates are used only to determine the selected point indices. The output point cloud retains the original Euclidean coordinates and associated attributes, including normals, and labels, so no geometric distortion is introduced into the downstream segmentation model.

The procedure requires only coordinate transformation, voxel indexing, hashing, and representative selection, yielding O(N) complexity. If surface normals are not provided, normal estimation typically adds $O(N\log N)$ preprocessing cost.

Note that the geometric interpretation is local rather than exact, and may become less accurate near sharp edges, depth discontinuities, or highly irregular surfaces. Even in such cases, the sampling procedure remains well-defined and stable.

## 4. Benchmark Protocol on SIP

### *4.1 Dataset Characteristics*

The SIP dataset [13] is a construction-focused benchmark of single-station terrestrial LiDAR scans designed for 3D segmentation, from which the indoor subset is used in this study. SIP preserves the raw characteristics of single-scan acquisition in active construction sites:

- Geometry-driven sampling bias (incidence and range) from a fixed sensor
- Severe occlusion and partial visibility in cluttered construction scenes
- Object-centric long-tail distribution, with dominant planar structures (wall, ceiling, floor) and sparse construction elements (pipes, columns, ladders, stairs)

SIP exhibits a strong long-tailed distribution, which is quantified using the statistics in Table 1. The dataset is highly imbalanced: dominant planar classes such as wall, ceiling, and floor account for the majority of points, while construction-specific objects remain extremely sparse. In particular, ladder (0.2%) and stair (0.9%) occupy only a negligible fraction of the dataset and appear in a limited number of scans.

This imbalance is further amplified at the point level, where rare classes contain orders of magnitude fewer points than dominant surfaces, making them highly sensitive to sampling and training dynamics.

**Table 1.** Class distribution and scan-level statistics of the SIP indoor subset. mCP is mean ± std. Others are ignored.

| Class | Global (%) | mCP (%) | Scan Presence (%) |
|---|---|---|---|
| wall | 34.3 | 37.0 ±13.8 | 100.0 (27/27) |
| ceiling | 30.9 | 30.0 ±9.1 | 96.3 (26/27) |
| floor | 23.1 | 22.4 ±6.8 | 100.0 (27/27) |
| pipes | 5.5 | 6.2 ±4.0 | 88.9 (24/27) |
| column | 1.9 | 2.6 ±1.2 | 66.7 (18/27) |
| ladder | 0.2 | 0.6 ±0.4 | 33.3 (9/27) |
| stair | 0.9 | 4.6 ±5.1 | 22.2 (6/27) |
| Others (ignored) | 3.1 | - | - |

To characterize the long-tailed distribution, both dataset-level and scan-level statistics are used. The global class proportion is defined as:

$$p_c = \frac{N_c}{\sum_k N_k}$$

At the scan level, the class proportion is:

$$p_{s,c} = \frac{N_{s,c}}{\sum_{k \in \mathcal{C}_{valid}} N_{s,k}}$$

Based on this, the mean class proportion (mCP) and presence frequency are defined as:

$$\mathrm{mCP}_c = \frac{1}{\mid \mathcal{S}_c \mid} \sum_{s \in \mathcal{S}_c} p_{s,c}\,, \qquad f_c = \frac{\mid \mathcal{S}_c \mid}{\mid \mathcal{S} \mid}$$

where $\mathcal{S}_c = \{s \mid N_{s,c} > 0\}$.

*4.2 Evaluation protocol*

From the original 23 annotated classes, 7 classes that appear in at least four scans in the indoor subset are selected: {wall, ceiling, floor, pipes, column, ladder, stair}. This ensures minimum representation across scans while retaining construction-relevant objects.

The dataset is split into 21 training and 6 testing scans. To ensure fair evaluation under long-tail conditions, rare classes are distributed across splits (Table 2). Frequent classes appear in nearly all scans across both splits.

Due to the limited number of scans and extreme class imbalance, constructing a separate validation set or applying k-fold cross-validation is not feasible without further reducing rare class coverage. Instead, experiments are conducted using multiple random seeds, and results are reported as mean ± standard deviation.

**Table 2.** Train/test distribution of rare classes to ensure balanced evaluation.

| Rare Class | Train (21 scans) | Test (6 scans) |
|---|---|---|
| ladder | 6 | 3 |
| stair | 3 | 3 |

Evaluation metrics follow standard practice in 3D semantic segmentation, including Overall Accuracy (allAcc), Mean Accuracy (mAcc), and Mean Intersection over Union (mIoU). As the method operates on downsampled point clouds, evaluation is performed on the reconstructed full-resolution point cloud via inverse mapping from sampled points, unless stated otherwise.

### *4.3 Input scale and scene fragmentation*

Large-scale point clouds require downsampling and spatial partitioning before model input. Under a fixed point budget, these operations jointly determine the geometric detail and spatial context available to the model: finer sampling preserves local structure but increases fragmentation, whereas coarser sampling expands fragment-level coverage at the cost of detail. This trade-off is particularly important for SIP, where thin and sparsely represented classes are sensitive to both point removal and fragmented context.

Accordingly, the training and inference pipeline was designed to maintain a fixed input budget while preserving full-scene coverage during evaluation, as summarized in Fig. 2. The following configurations were adopted:

- *Input and geometry preservation*. Each point uses xyz+normal features, and no center shift is applied, preserving the sensor-centered radial geometry.
- *Fragmentation strategy (Algorithm 2).* Scene fragments are generated using cylindrical cropping, rather than standard spherical cropping, preserving vertical spatial relationships. Full-scene coverage is maintained, with overlapping regions aggregated via stitching during evaluation.
- *Imbalance-aware training.* Focal loss and Lovász loss [42] are adopted, together with rare-class-anchored fragments, to reduce suppression of rare-class gradients.

**Algorithm 2.** Scene fragmentation.

**Input:** point cloud $X$, labels $Y$, point budget $K$, mode $\in$ {train, test}
**Output:** fragment set $\mathcal{F}$

1: $N \leftarrow \mid X \mid$
2: **if** $N \leq K$ **then return** $\{X\}$
3: $M \leftarrow \left\lceil \frac{N}{K} \right\rceil$
4: $\mathcal{F} \leftarrow \emptyset$, $covered \leftarrow \mathbf{0}_N$
5: // Fragment Sampling
6: **for** $i = 1 \ldots M$ **do**
7: $a \leftarrow$ random anchor not in $covered$
8: $I \leftarrow cylinderCrop(a, K)$
9: $\mathcal{F} \leftarrow \mathcal{F} \cup \{I\}$
10: $covered[I] \leftarrow 1$
11: // Rare-Class Anchoring
12: **if** mode = train **and** $\mathcal{C}_r \neq \emptyset$ **then**
13: **for each** class $c \in \mathcal{C}_r$ **do**
14: **while** insufficient fragments for $c$ **do**
15: $a \leftarrow$ random point with label $c$
16: $I \leftarrow \text{kNN}(a, K)$
17: $\mathcal{F} \leftarrow \mathcal{F} \cup \{I\}$
18: // Coverage Completion
19: **if** mode = test **then**
20: **while** $covered \neq \mathbf{1}_N$ **do**
21: $a \leftarrow$ random uncovered point
22: $I \leftarrow cylinderCrop(a, K)$
23: $\mathcal{F} \leftarrow \mathcal{F} \cup \{I\}$
24: $covered[I] \leftarrow 1$
25: **return** $\mathcal{F}$

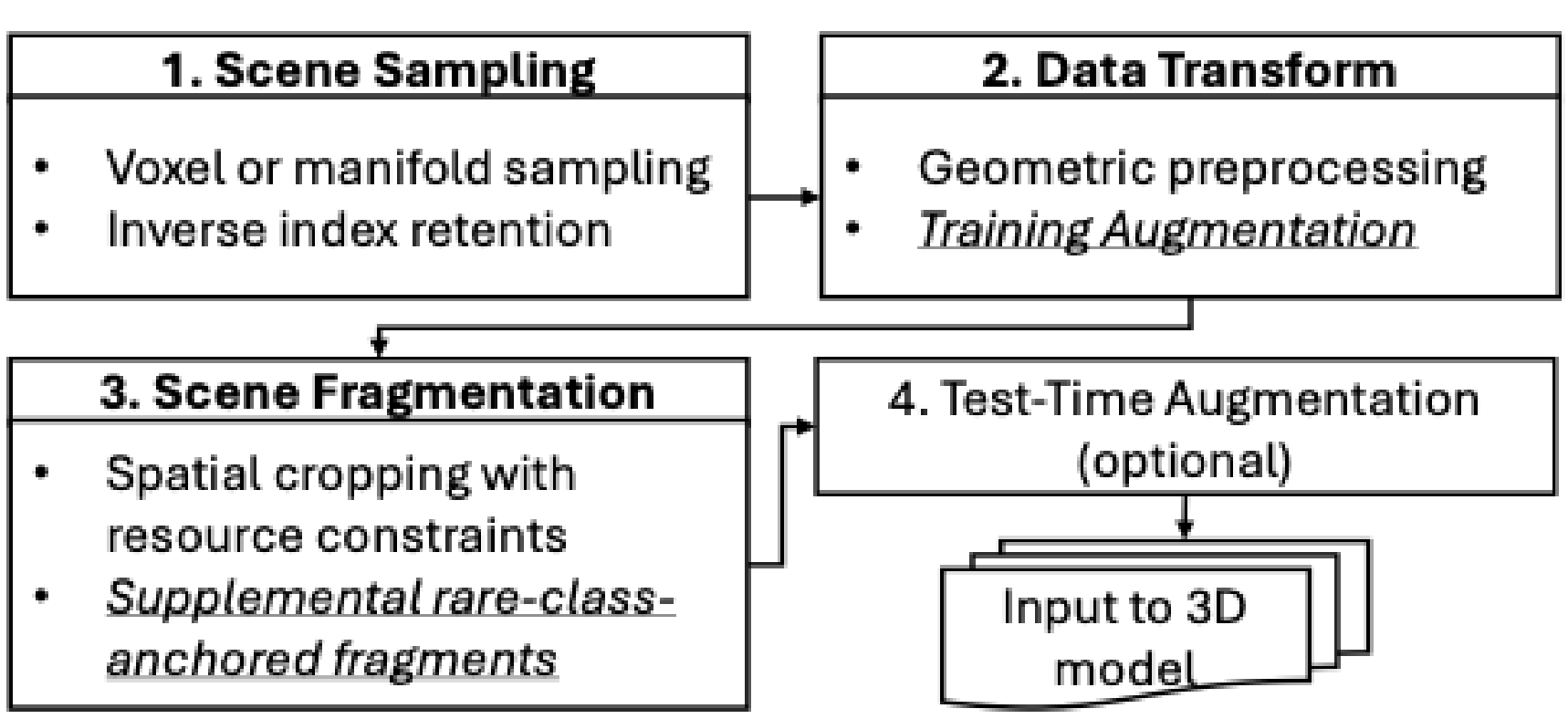

**Fig. 2.** Training and inference pipeline for SIP. Underlined text indicates training-only components.

5. Experimental Results

*5.1. Design considerations*

The maximum number of points that can be processed in a scene fragment defines the primary computational constraint in large-scale 3D segmentation. Accordingly, all configurations are evaluated under the same per-fragment point budget. Sampling resolution and point allocation are not independently normalized beyond this constraint, because they determine how the available input capacity is translated into spatial context and geometric evidence. The experimental design therefore examines two sampling-dependent outcomes:

- *Fragment extent as an outcome.* The spatial extent represented by each fragment is allowed to vary with the sampling resolution and retained point distribution, rather than being explicitly matched across configurations.
- *Surface-wise point allocation.* Euclidean grid sampling and manifold sampling differ in how the available points are distributed across the observed surfaces.

All experiments use the same maximum number of points per fragment and follow a common scan-level training protocol across resolutions. Fragment losses are aggregated before each scan-level update to maintain a consistent epoch definition despite differences in fragment count. Detailed training settings are summarized in Table 3. All experiments were conducted on a single NVIDIA GeForce RTX 3090 GPU with 24 GB VRAM.

**Table 3.** Training protocol and backbone configurations.

| Category | Setting |
|---|---|
| *Training protocol* | |
| Update scheme | Scan-level updates with fragment-loss accumulation; fragment batch size = 4 |
| Training length | 80 epochs |
| Input budget | Maximum points per fragment = 30K |
| Repeats | 5 runs per configuration (mean ± std) |
| *Optimization setting* | |
| Loss | Focal ($\gamma$=1.0) + 0.5×Lovasz-Softmax |
| Optimizer | AdamW (lr=0.001, weight_decay=0.01) |
| Scheduler | MultiStepLR (milestones=[0.6, 0.8], $\gamma$ =0.1) |
| *Backbone Configuration* | |
| PT | Encoder depths: (1,2,3,5,2), channels: (32,64,128,256,512), strides: (1,4,4,4,4), neighbors: (8,16,16,16,16) |

| PNxt | Stem width: 32; encoder channels: (64, 128, 256, 512); InvResMLP blocks: (1, 2, 1, 1) |
|---|---|

Point Transformer (PT) [33] and PointNeXt (PNxt) [43] are used as the primary backbones. PT follows the Pointcept implementation [44], whereas PNxt-B was integrated into the same training and evaluation pipeline for a controlled comparison. Both are hierarchical point-based models that construct local neighborhoods directly from point coordinates, avoiding dependence on a voxel lattice. For PT, an earlier architecture without the explicit partitioning or serialization mechanisms used in later variants [45], [46] is adopted to reduce interaction between the backbone and the input sampling scheme. Together, the two backbones allow the sampling effect to be examined across different feature-aggregation mechanisms within the hierarchical point-based model family.

### *5.2. Resolution-wise Sampling Performance*

Sampling resolution controls the trade-off between local geometric detail and fragment-level spatial support under the fixed input budget. Finer sampling retains denser geometric evidence but increases fragmentation, whereas coarser sampling reduces the number of fragments and expands the spatial extent represented in each forward pass, as illustrated in Figure 3. Figure 4 further shows the retained point count and fragment count across the five tested resolutions under the 30K-point limit per fragment. Based on this setting, Euclidean grid sampling and manifold sampling are evaluated at resolutions of 0.04, 0.06, 0.09, 0.13, and 0.18 m using both PT and PNxt.

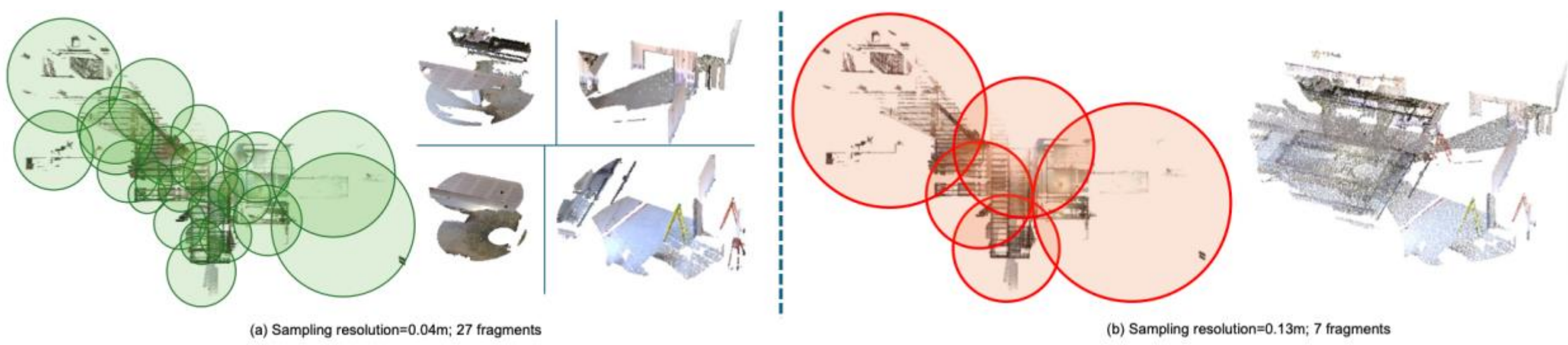

(a) Sampling resolution=0.04m; 27 fragments

(b) Sampling resolution=0.13m; 7 fragments

**Fig. 3.** Fragmentation patterns under different sampling resolutions: (a) 0.04 m with 27 fragments and (b) 0.13 m with 7 fragments. Representative fragment extents are shown beside each scene.

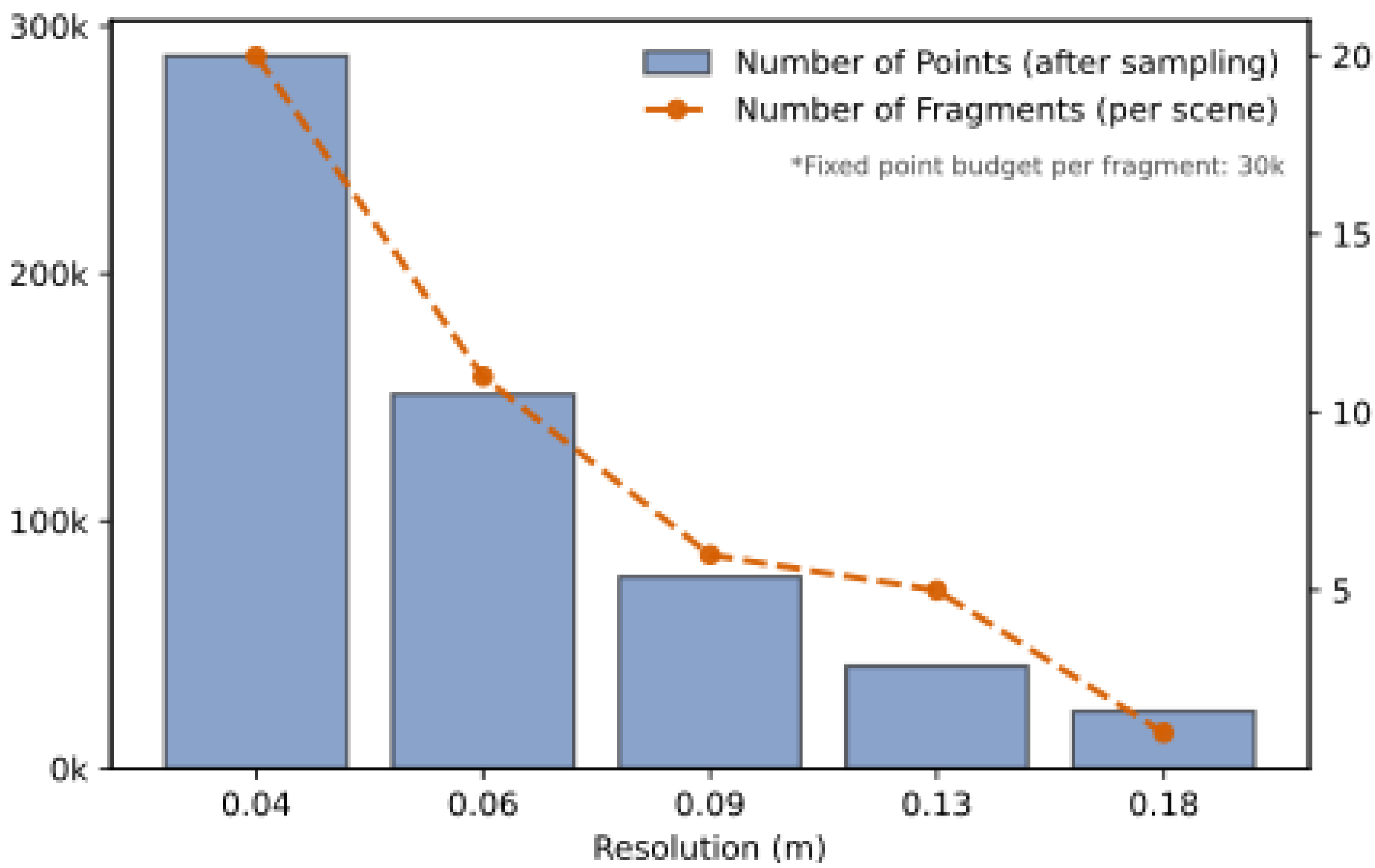

**Fig. 4.** Retained point count and fragment count across sampling resolutions under the 30K-point limit per fragment.

Tables 4 and 5 report the full class-wise IoU results and aggregate metrics for PT and PNxt, respectively. Figure 5 summarizes the corresponding mIoU and NP-IoU trends across sampling resolutions. Here, NP denotes the non-planar construction elements—pipes, column, ladder, and stair. For PT, manifold sampling improves mIoU and NP-IoU at most resolutions and substantially reduces the degradation observed with grid sampling at the coarsest setting. PNxt also shows clear gains at fine-to-moderate resolutions, particularly from r04 to r09.

**Table 4.** PT segmentation results across sampling resolutions using grid and manifold sampling.

| | PT-Base | | | | | PT-Manifold | | | | |
|---|---|---|---|---|---|---|---|---|---|---|
| | r04 | r06 | r09 | r13 | r18 | r04 | r06 | r09 | r13 | r18 |
| wall | 83.6 ± 0.9 | 87.3 ± 1.5 | 89.1 ± 0.7 | 88.9 ± 1.0 | 71.7 ± 15.3 | 79.4 ± 4.6 | 88.5 ± 0.6 | 87.9 ± 0.9 | 87.8 ± 1.0 | 85.4 ± 1.8 |
| ceiling | 86.3 ± 1.8 | 90.6 ± 1.3 | 91.0 ± 0.7 | 89.7 ± 1.2 | 79.3 ± 7.3 | 86.0 ± 2.5 | 90.6 ± 0.6 | 89.6 ± 1.2 | 88.4 ± 0.6 | 87.6 ± 2.0 |
| floor | 91.4 ± 1.0 | 93.3 ± 2.5 | 93.0 ± 2.8 | 96.8 ± 0.7 | 87.3 ± 8.1 | 88.6 ± 4.8 | 94.8 ± 0.5 | 95.4 ± 0.2 | 96.0 ± 0.6 | 95.7 ± 1.5 |
| pipes | 52.1 ± 12.9 | 67.1 ± 9.4 | 71.0 ± 4.1 | 61.8 ± 5.3 | 43.7 ± 15.6 | 68.0 ± 3.1 | 70.5 ± 0.6 | 64.5 ± 4.6 | 60.3 ± 2.6 | 57.8 ± 5.0 |
| column | 32.6 ± 9.1 | 42.0 ± 8.4 | 45.6 ± 11.1 | 63.4 ± 2.9 | 27.5 ± 22.1 | 37.5 ± 4.1 | 51.0 ± 6.5 | 44.7 ± 12.4 | 46.3 ± 12.1 | 41.4 ± 12.2 |
| ladder | 27.9 ± 19.9 | 60.6 ± 15.8 | 49.1 ± 7.7 | 50.1 ± 7.1 | 19.9 ± 11.0 | 66.2 ± 2.3 | 71.7 ± 9.7 | 62.7 ± 15.8 | 50.2 ± 7.7 | 38.4 ± 5.2 |
| stair | 20.1 ± 9.5 | 34.9 ± 9.1 | 42.1 ± 11.4 | 63.3 ± 9.5 | 44.9 ± 19.5 | 34.8 ± 8.8 | 41.1 ± 3.3 | 49.1 ± 0.9 | 52.6 ± 8.3 | 55.0 ± 18.6 |
| mIOU | 56.3 ± 4.2 | 68.0 ± 3.7 | 68.7 ± 2.9 | 73.4 ± 1.4 | 53.5 ± 11.9 | 65.8 ± 4.1 | 72.6 ± 2.7 | 70.5 ± 2.4 | 68.8 ± 1.7 | 65.9 ± 1.2 |
| NP-IOU | 33.2 ± 7.1 | 51.2 ± 5.7 | 54.1 ± 4.6 | 58.4 ± 2.4 | 36.1 ± 13.6 | 51.6 ± 4.2 | 58.5 ± 4.3 | 58.8 ± 4.1 | 54.4 ± 3.0 | 50.4 ± 1.9 |

**Table 5.** PNxt results across sampling resolutions using grid and manifold sampling.

| | PNxt-Base | | | | | PNxt-Manifold | | | | |
|---|---|---|---|---|---|---|---|---|---|---|
| | r04 | r06 | r09 | r13 | r18 | r04 | r06 | r09 | r13 | r18 |
| wall | 68.5 ± 3.2 | 70.7 ± 3.4 | 64.9 ± 8.0 | 68.0 ± 6.4 | 60.3 ±10.3 | 68.8 ± 1.7 | 66.0 ± 3.9 | 70.0 ± 3.8 | 69.5 ± 5.7 | 65.4 ± 1.8 |
| ceiling | 59.5 ± 15.2 | 80.5 ± 2.5 | 81.9 ± 0.9 | 80.5 ± 5.0 | 84.9 ± 1.0 | 59.8 ± 30.6 | 82.3 ± 1.2 | 82.7 ± 0.7 | 68.2 ± 18.4 | 79.7 ± 4.4 |
| floor | 85.9 ± 4.8 | 88.3 ± 3.3 | 85.3 ± 0.5 | 87.8 ± 3.9 | 91.8 ± 1.5 | 83.5 ± 1.9 | 86.6 ± 2.3 | 91.6 ± 2.2 | 93.3 ± 0.2 | 92.3 ± 0.5 |
| pipes | 16.6 ± 7.7 | 45.5 ± 9.5 | 46.6 ± 4.0 | 43.9 ± 8.2 | 49.4 ± 3.1 | 31.6 ± 16.7 | 52.8 ± 1.6 | 55.1 ± 3.0 | 27.6 ± 14.5 | 38.1 ± 7.3 |
| column | 7.5 ± 1.5 | 4.7 ± 3.5 | 6.7 ± 1.6 | 7.6 ± 0.6 | 7.2 ± 0.7 | 2.8 ± 0.1 | 9.2 ± 0.4 | 9.5 ± 2.0 | 8.8 ± 0.2 | 7.9 ± 1.3 |
| ladder | 0.1 ± 0.1 | 3.1 ± 2.7 | 5.9 ± 5.3 | 13.4 ± 5.9 | 6.2 ± 2.4 | 19.6 ± 8.1 | 30.5 ± 5.3 | 23.9 ± 5.6 | 24.6 ± 10.2 | 14.6 ± 2.8 |
| stair | 4.9 ± 1.5 | 2.7 ± 1.0 | 2.6 ± 2.6 | 6.0 ± 2.3 | 17.4 ± 1.3 | 5.1 ± 0.1 | 8.8 ± 4.0 | 11.4 ± 0.5 | 13.1 ± 0.6 | 9.7 ± 5.0 |
| mIOU | 34.7 ± 4.9 | 42.2 ± 0.9 | 42.0 ± 1.5 | 43.9 ± 0.5 | 45.3 ± 1.3 | 38.8 ± 7.9 | 48.0 ± 1.9 | 49.2 ± 0.5 | 43.6 ± 4.1 | 44.0 ± 2.5 |
| NP-IOU | 7.3 ± 2.7 | 14.0 ± 0.9 | 18.4 ± 0.8 | 21.1 ± 0.4 | 24.3 ± 0.6 | 14.8 ± 6.2 | 25.3 ± 2.5 | 30.1 ± 0.4 | 21.8 ± 1.8 | 20.8 ± 3.5 |

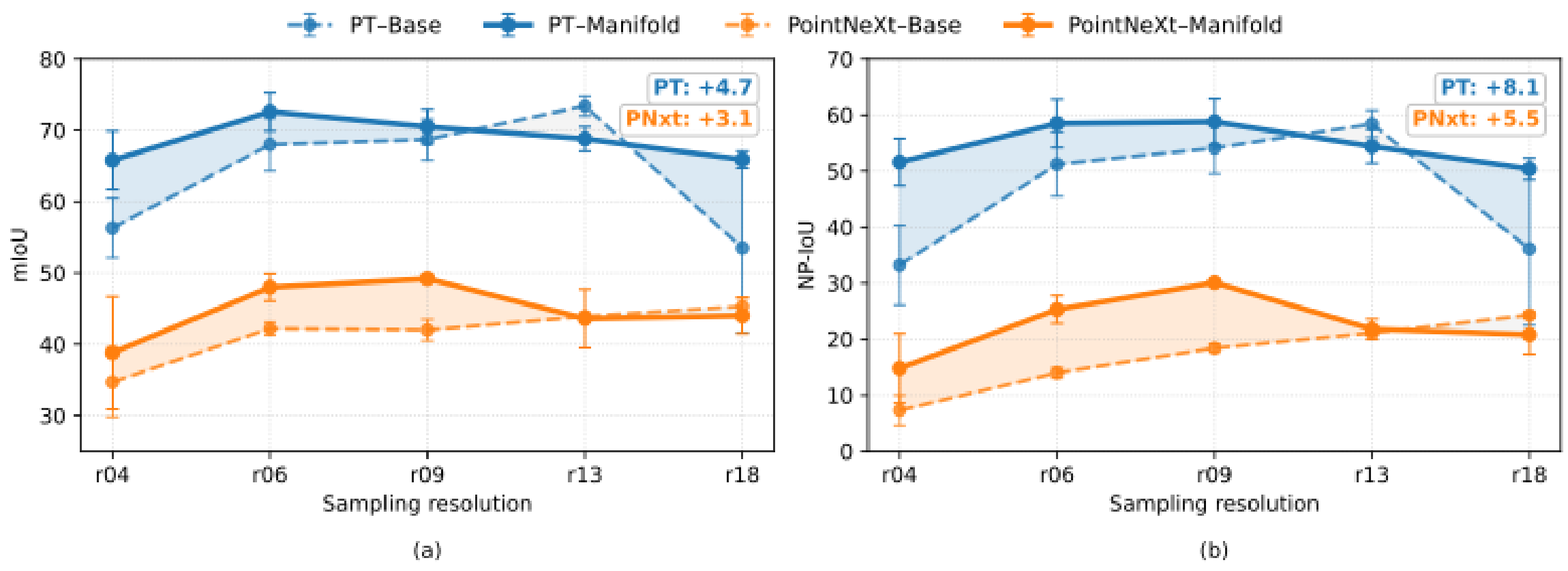

**Fig. 5.** Segmentation performance across sampling resolutions for PT and PNxt: (a) mIoU and (b) NP-IoU. Labels indicate the average gain of manifold sampling over grid sampling.

Table 6 summarizes the performance averaged across the five sampling resolutions. Manifold sampling improves all reported metrics for both PNxt and PT, with larger gains observed in mIoU and the NP-specific measures. For PNxt, the average mIoU increases from 41.6 to 44.7 and NP-IoU from 17.0 to 22.6. For PT, the corresponding gains are from 64.0 to 68.7 in mIoU and from 46.6 to 54.7 in NP-IoU. The accuracy-based metrics show the same overall trend.

**Table 6.** Resolution-averaged performance of PT and PNxt across the five sampling resolutions. Parenthetical values indicate changes from grid to manifold sampling.

| metric | PNxt-Base | PNxt-Manifold | PT-Base | PT-Manifold |
|---|---|---|---|---|
| mIoU | 41.6 | **44.7 (+3.1)** | 64.0 | **68.7 (+4.7)** |
| NP-IoU | 17.0 | **22.6 (+5.5)** | 46.6 | **54.7 (+8.1)** |
| mAcc | 52.4 | **56.4 (+4.0)** | 73.1 | **77.7 (+4.6)** |
| NP-Acc | 29.8 | **37.1 (+7.2)** | 58.2 | **66.9 (+8.7)** |
| AllAcc | 79.7 | 79.8 (+0.2) | 90.1 | **91.9 (+1.8)** |

Among the NP classes, the largest improvement is observed for ladder. The resolution-averaged ladder IoU increases from 41.5 to 57.8, a gain of 16.3 points for PT and from 5.7 to 22.6, a gain of 16.9 points for PNxt. This consistent gain across the two models indicates that manifold sampling is particularly effective in preserving the limited geometric evidence associated with sparsely represented elements.

Across both sampling strategies, PT consistently outperforms PNxt in the aggregate metrics. This difference may reflect PT's attention-based feature aggregation, which can more effectively exploit local geometric relationships in partially observed scenes. More importantly, both models benefit from manifold sampling on average, and exhibit more stable behavior across sampling resolutions. These results suggest that the proposed sampling improves the quality of the geometric representation presented to the models.

### *5.3. Qualitative Analysis*

Figure 6 compares grid and manifold sampling using the PT run closest to the mean mIoU. Both methods remain largely consistent on dominant planar regions, while manifold sampling provides clearer and more continuous predictions for thin, sparse, and geometrically complex elements.

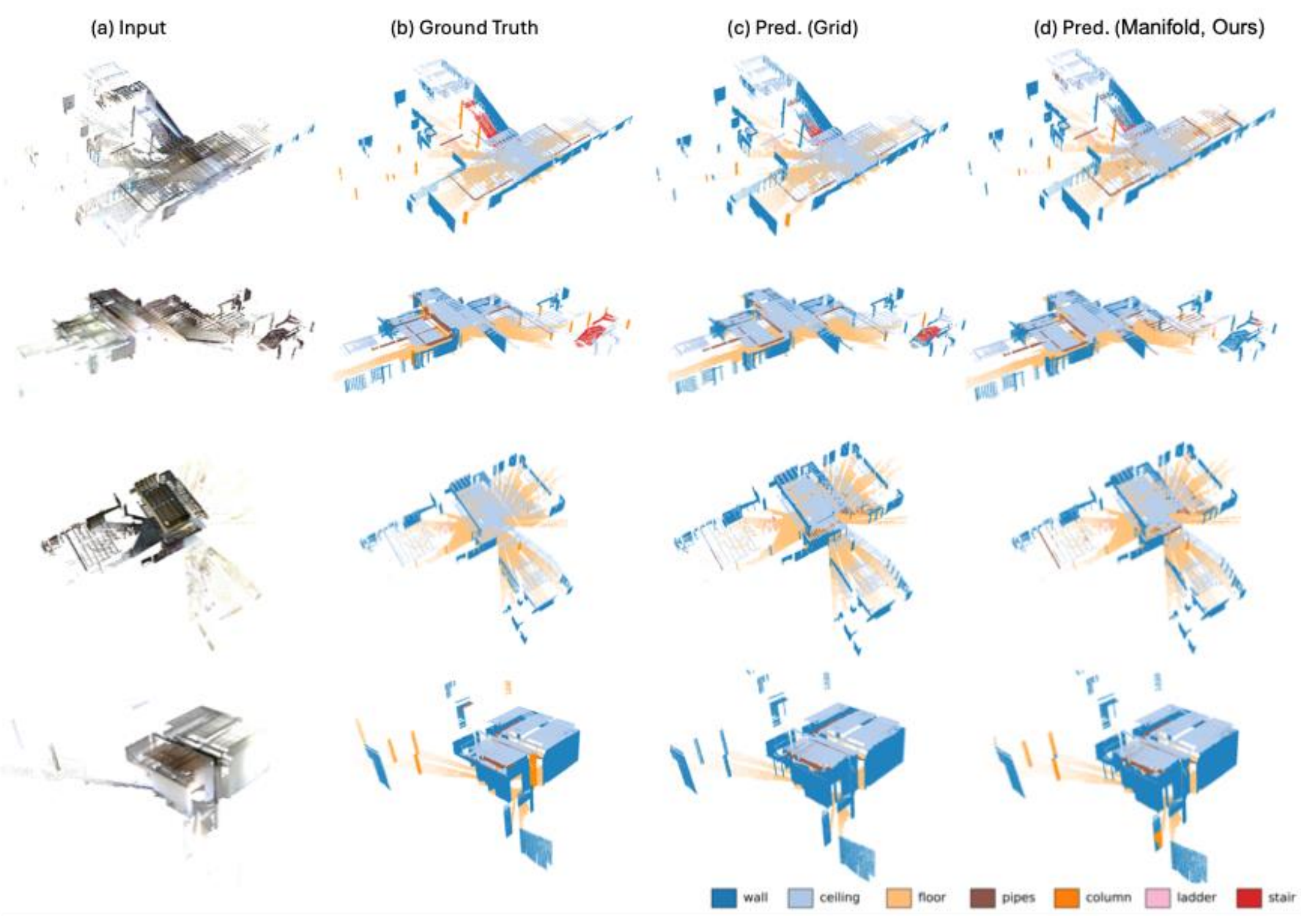

**Fig 6.** Qualitative comparison of segmentation results. From left to right are the input point cloud, ground truth, prediction with grid sampling, and prediction with Manifold (ours).

Figures 7 and 8 show representative success and failure cases. Manifold sampling improves the recovery of ladders and pipe-like elements, but stair railings remain challenging because their thin and partial geometry resembles pipes or ladders. Some errors also suggest reliance on height-related context: in multi-level scenes, railings at elevations similar to pipe installations on another floor may be classified as pipes. Other failures include column–wall confusion and stair railings classified as ladders, indicating that manifold sampling cannot fully resolve ambiguity among classes with similar local geometry and spatial context.

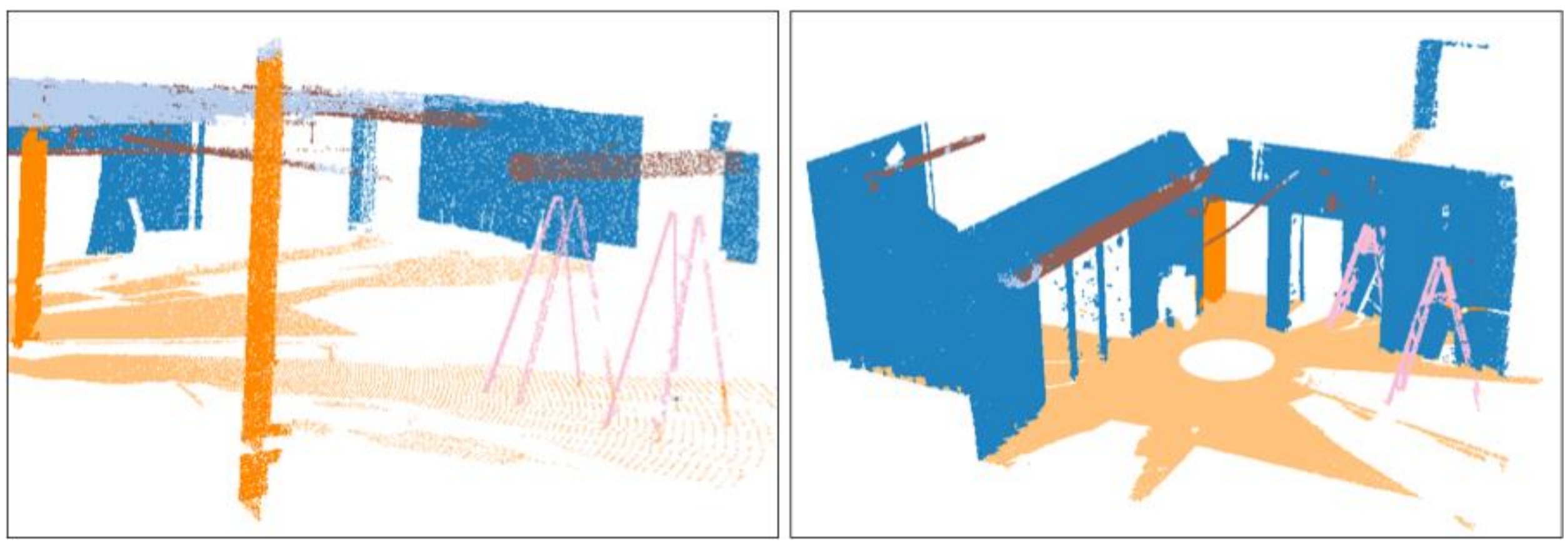

**Fig 7.** Success cases of Manifold. Incidence-aware allocation helps recover thin and sparse elements such as ladders and pipes.

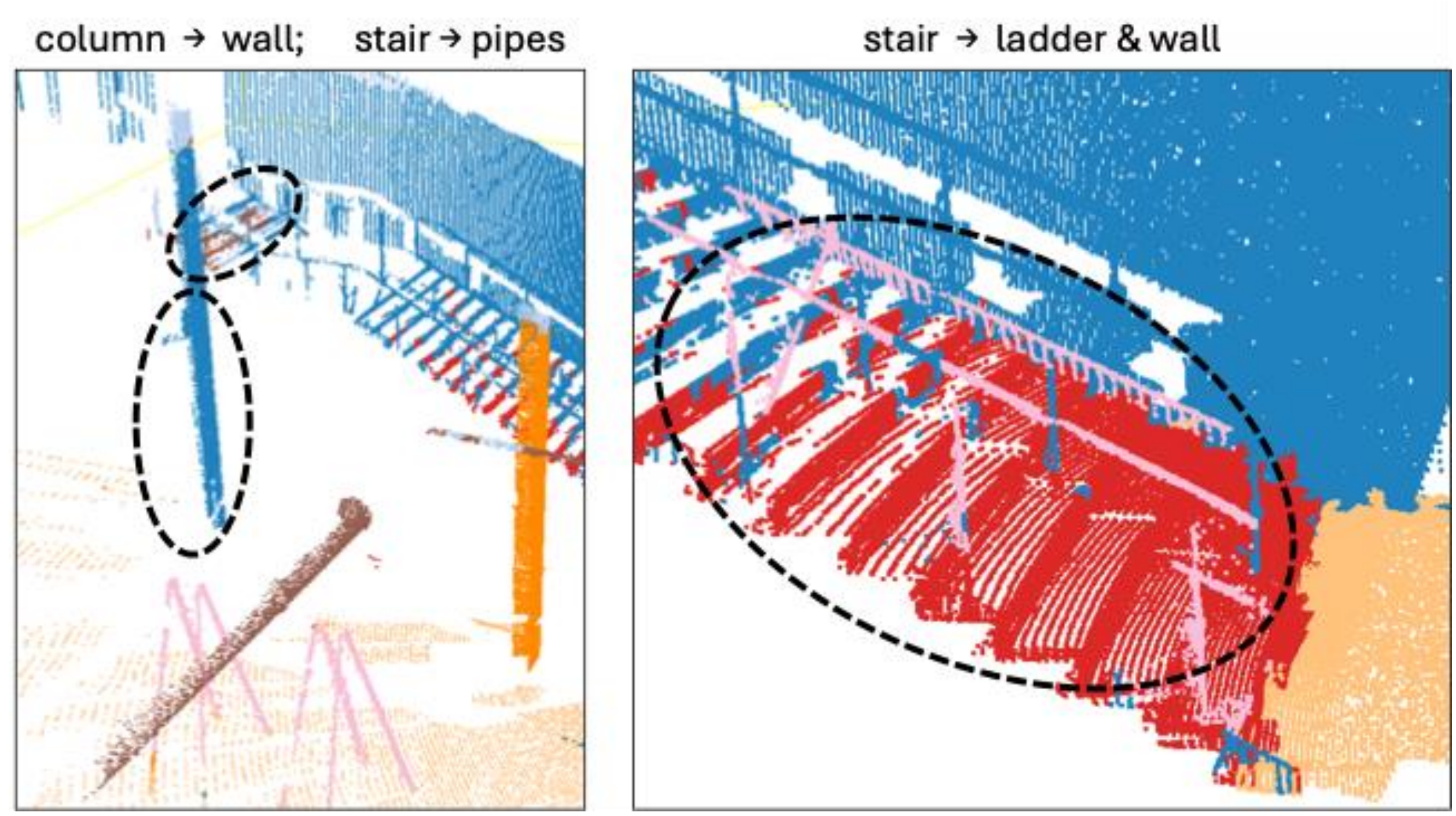

**Fig 8.** Failure cases of Manifold. Structurally similar elements remain confused when local geometric cues overlap.

## 6. DISCUSSION

### *6.1. Reference comparison with FPS*

Figure 9 compares the retained-point distributions produced by grid sampling, farthest point sampling (FPS), manifold sampling, and manifold+. Manifold+ combines Euclidean grid samples with manifold-selected points to preserve lattice coverage while incorporating incidence-aware allocation. FPS promotes broad Euclidean separation, but its iterative distance computation makes it less suitable for repeated use in large-scale scene training.

Table 7(a) reports the corresponding PT results at 0.09 m. Because FPS does not provide a direct inverse mapping to the original scan, its performance is evaluated only on the sampled points and should therefore be interpreted separately from the other methods. Under this setting, FPS provides competitive mIoU but does not improve NP-IoU over the grid baseline. Manifold sampling improves both metrics, while manifold+ achieves the highest mIoU and NP-IoU in this reference comparison.

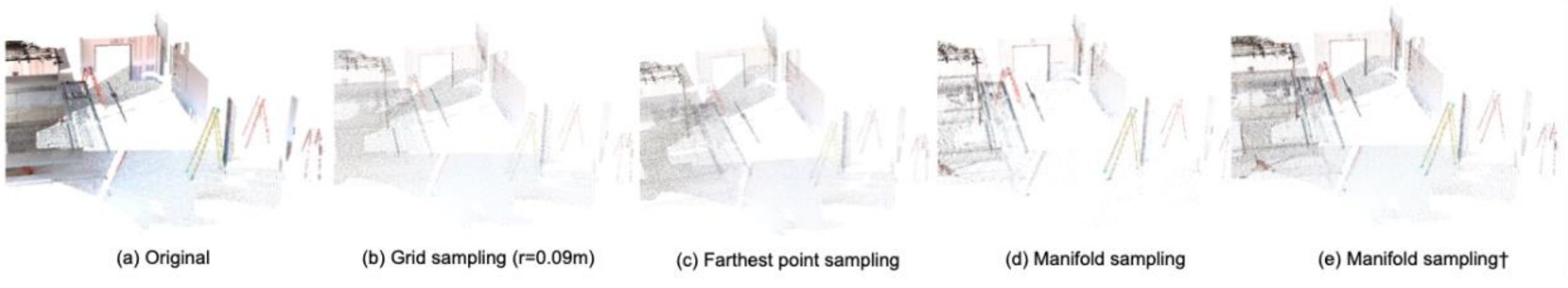

**Fig 9.** Sampling patterns under the same nominal resolution. Manifold emphasizes incidence-aware geometric cues, while manifold+ combines this allocation with Euclidean grid coverage.

**Table 7.** Reference sampling comparison at 0.09 m for (a) PT and (b) SpUNet.

| (a) PT | Base | FPS | Manifold | Manifold+ |
|---|---|---|---|---|
| mIoU | 68.7 | 70.0 | 70.5 | **72.7** |
| NP-IoU | 54.1 | 53.9 | 58.8 | **60.2** |

| (b) SpUNet | Base | Manifold | Manifold+ |
|---|---|---|---|
| mIoU | **59.6** | 51.8 | 56.5 |
| NP-IoU | 40.3 | 28.2 | **40.4** |

*FPS is evaluated on sampled points only; the remaining methods are evaluated after inverse mapping to the original scan.

### *6.2. Architecture-specific behavior*

Table 7(b) presents the corresponding comparison using SpUNet [47] at 0.09 m. Unlike the point-based backbones, SpUNet relies on sparse convolution over a Euclidean voxel lattice. Pure manifold sampling reduces both mIoU and NP-IoU, suggesting that the altered Euclidean occupancy is less compatible with the neighborhood structure expected by the backbone.

Manifold+ partially mitigates this limitation by retaining the grid-sampled points while adding manifold-selected points. It recovers NP-IoU from 28.2 to 40.4, approximately matching the grid baseline of 40.3. Manifold+ therefore offers a practical lattice-preserving variant for applying incidence-aware sampling to voxel-based backbones.

## 7. CONCLUSION

This study reframed individual construction LiDAR scans as local site evidence for scene understanding and examined the effects of sampling design under realistic sensing and input constraints. To address acquisition-driven density bias, this study introduced incidence-aware manifold sampling as a lightweight front-end normalization strategy that redistributes point density according to surface geometry. Experiments on the SIP benchmark showed that the proposed method improves resolution-averaged segmentation performance across two hierarchical point-based models, with particularly strong gains for non-planar construction elements such as ladders. The method also reduced sensitivity to sampling resolution, indicating that the geometric representation presented to the models remains more stable across operating scales. These findings suggest that sampling should be treated as an active component of single-scan 3D segmentation rather than as a generic preprocessing step.

The proposed approach also supports engineering systems that must process large point clouds under limited sensing and computing capacity. More effective use of a fixed point budget can benefit robotic perception and frequent local updates of digital twins without requiring complete multi-view reconstruction at every cycle. Its low front-end overhead also makes it suitable for edge or low-resource deployment, while the observed incidence dependence highlights the importance of considering sensor placement together with downstream sampling.

Future work should focus on improving robustness to emerging challenges in single-scan 3D perception, including domain shift, limited task-specific supervision, transfer of pretrained representations, and variation in backbone architecture. Further investigation is also needed into architecture-aware sampling strategies and model generalization beyond dataset-specific operating points. Developing sampling and representation methods that remain effective under such variability remains an important direction for practical 3D perception.

## DATA AVAILABILITY STATEMENT

The data used in this study are available through the SIP benchmark at Zenodo, DOI: 10.5281/zenodo.17667735. The code and training configurations used in this study are available at GitHub: https://github.com/syoi92/Pointcept-SIP/. Additional trained model artifacts are available from the corresponding author upon reasonable request.